\documentclass{article}

\usepackage{microtype}
\usepackage{graphicx}
\usepackage{subcaption}
\usepackage{booktabs} %

\usepackage{hyperref}
\usepackage{adjustbox}
\usepackage{wrapfig}
\usepackage{stfloats}

\usepackage[preprint]{icml2026}

\usepackage{amsmath}
\usepackage{amssymb}
\usepackage{mathtools}
\usepackage{amsthm}
\usepackage[breakable]{tcolorbox}

\usepackage[ruled]{algorithm2e}
\usepackage{algorithm}
\usepackage{float}
\usepackage{bold-extra}
\usepackage{multirow}
\usepackage{makecell}

\usepackage[capitalize,noabbrev]{cleveref}

\theoremstyle{plain}

\theoremstyle{definition}

\theoremstyle{remark}

\usepackage[textsize=tiny]{todonotes}

\newcommand{\M}{\mathbf{M}}
 
\newcommand{\w}{\mathrm{w}}

\icmltitlerunning{ToolMol: Evolutionary Agentic Framework for Multi-objective  Drug Discovery}

\begin{document}

\twocolumn[
  \icmltitle{ToolMol: Evolutionary Agentic Framework for Multi-objective  Drug Discovery}

  \icmlsetsymbol{equal}{*}

  \begin{icmlauthorlist}
    \icmlauthor{Andrew Y. Zhou}{cse}
    \icmlauthor{Sharvaree Vadgama}{cse}
    \icmlauthor{Sumanth Varambally}{cse}
    \icmlauthor{Peter Eckmann}{stanford}\\
    \icmlauthor{Michael K. Gilson}{skaggs}
    \icmlauthor{Rose Yu}{cse}
  \end{icmlauthorlist}

  \icmlaffiliation{cse}{Department of Computer Science and Engineering, UC San Diego, La Jolla, California, United States}
  \icmlaffiliation{skaggs}{Skaggs School of Pharmacy and Pharmaceutical Sciences, UC San Diego, La Jolla, California, United States}
  \icmlaffiliation{stanford}{Department of Computer Science, Stanford University, Stanford, California}

  \icmlcorrespondingauthor{Andrew Y. Zhou}{anz008@ucsd.edu}
  \icmlcorrespondingauthor{Rose Yu}{roseyu@ucsd.edu}

  \icmlkeywords{Machine Learning, ICML}

  \vskip 0.3in
]

\printAffiliationsAndNotice{}  %

\begin{abstract}
Advances in large language models (LLMs) have recently opened new and promising avenues for small-molecule drug discovery. Yet existing LLM-based approaches for molecular generation often suffer from high rates of invalid and low-quality ligand candidates, a result of the syntactic limitations of current models with regard to molecular strings. In this paper, we introduce \texttt{ToolMol}, an evolutionary agentic framework for de novo drug design. \texttt{ToolMol} combines a multi-objective genetic algorithm with an agentic LLM operator that iteratively updates the ligand population. We build a comprehensive toolbox of RDKit-backed functions that allows our agentic operator to consisently make precise ligand modifications. \texttt{ToolMol} achieves state-of-the-art performance on multi-objective property optimization tasks, discovering drug-like and synthesizable ligands that have $>10\%$ stronger predicted binding affinity compared to existing methods, evaluated on three protein targets. $\texttt{ToolMol}$ ligands additionally achieve state-of-the-art results in gold-standard Absolute Binding Free Energy scores, gaining over existing methods by over $35\%$. By studying chain-of-thought reasoning traces, we observe that tool-calling enables the model to more faithfully execute its planned modifications, efficiently exploiting the strong chemical prior knowledge in LLMs. 
\end{abstract}

\section{Introduction}
Small molecule drug discovery is a resource-intensive process that requires generated compounds to satisfy many crucial properties, historically requiring many rounds of wet lab trial-and-error. Advances in machine learning have yielded many generative methods that aim to solve this problem. Most previous work has focused on specialized generative models such as VAEs \citep{eckmann2022limolatentinceptionismtargeted, eckmann2025mflaldrugcompoundgeneration, pmlr-v162-noh22a}, diffusion models \citep{lee2023exploringchemicalspacescorebased,  zhou2024decompoptcontrollabledecomposeddiffusion, joshi2025allatomdiffusiontransformersunified, guan2024decompdiffdiffusionmodelsdecomposed, dorna2024tagmoltargetawaregradientguidedmolecule}, and group equivariant diffusion models \citep{hoogeboom2022equivariantdiffusionmoleculegeneration, vadgama2026probing, liu2025cliffordgroupequivariantdiffusion}. Optimization frameworks such as evolutionary algorithms \citep{Jensen2019} and Bayesian optimization algorithms \citep{zhu2023sampleefficientmultiobjectivemolecularoptimization} have also been applied to this problem. However, existing generative methods often do not target important molecular properties, limiting their ability to generate ligands that simultaneously achieve desirable binding affinity, drug-likeness and synthesizability \citep{crucitti2024denovo}.

Large Language Models (LLMs) have recently begun to garner interest as a method to generate small molecules, showing promise in generating strong, drug-like ligands \citep{wang2025efficientevolutionarysearchchemical}. Unlike specialized generative models, LLMs benefit from large-scale pretraining on domain-relevant, scientific text, which gives them the distinct advantage of inherent familiarity with the optimization task, as well as with the practices and heuristics of chemical research (e.g. common reactions and lead optimization techniques)\citep{White2023}. Extensive LLM-related works have demonstrated the ability of current LLMs to predict molecular properties 
\citep{guo2023largelanguagemodelschemistry} and generate novel structures \citep{flamshepherd2023languagemodelsgeneratemolecules}. Most recently, MOLLEO \citep{wang2025efficientevolutionarysearchchemical} proposed a genetic algorithm that directly incorporates LLMs as a mutation and crossover operator to generate molecular offspring, outperforming many specialized generative models in generating ligands with desirable properties.

However, a significant drawback with current LLMs is that they often fail at generating syntactically valid molecular strings, even when prompted to inspect their outputs carefully. We observe that this failure occurs consistently, appearing in more than $30\%$ of attempted molecule generations on average, even on strong reasoning models like GPT-OSS-120B \citep{openai2025gptoss120bgptoss20bmodel}. This significantly hinders the progress of current LLM-based methods. For instance, MOLLEO falls back on weaker non-LLM based, deterministic crossover/mutation operators when the LLM operator fails to generate a valid SMILES result. We claim that this method of allowing an LLM to directly output molecular strings is an imperfect method of utilizing LLMs for efficient, property-based drug discovery.

In this work, we introduce \texttt{ToolMol}, a novel drug discovery algorithm focused on de novo small molecule drug design. ToolMol combines a multi-objective genetic algorithm with an agentic LLM operator that iterates upon the ligand population using a set of deterministic, RDKit-backed tools. ToolMol solves the problem of invalid molecular generations by exploiting the highly-optimized LLM tool-calling functionality prevalent in current models. Instead of allowing the LLM to modify the molecular string encoding directly, ToolMol abstracts this process by providing the LLM with tools that allow it to simply provide structural parameters for its desired modifications. This not only greatly decreases the number of invalid SMILES strings generated by the LLM, but also yields a significant improvement in the molecular properties of generated ligands, including predicted binding affinity, drug-likeness, and synthesizability.

We summarize the contributions of this work as below:

\begin{itemize}
     \item We present \texttt{ToolMol}, an evolutionary agentic drug discovery framework that combines a multi-objective genetic algorithm with an agentic LLM operator to consistently generate syntactically valid and property-optimizing molecules.
     \item We achieve state-of-the-art results in multi-objective property optimization across three protein targets, with predicted binding affinity gains exceeding $10\%$ over prior methods, as well as state-of-the-art Absolute Binding Free Energy scores for two studied targets, demonstrating the practical utility of LLMs for de novo drug design.
     \item We study the effectiveness of our framework through case studies, and observe that the agentic tool-calling process significantly improves concordance between the LLM's reasoning trace and the actual ligand modifications.
\end{itemize}

\section{Related Work}
\paragraph{Generative models for molecular design}

A variety of generative architectures have been developed for molecular design, each learning an implicit distribution over chemical space and leveraging an external oracle to guide generation toward molecules with desirable binding properties. VAE-based approaches such as \citep{eckmann2022limolatentinceptionismtargeted, jin2018junction, eckmann2025mflaldrugcompoundgeneration, gomez2018automatic} have shown promise, but are generally unaware of the 3D protein structure. To address this, DecompOpt \citep{zhou2024decompoptcontrollabledecomposeddiffusion} and DecompDiff \citep{guan2024decompdiffdiffusionmodelsdecomposed} are diffusion models that condition on the protein structure, and are further guided toward optimal ligand molecules by an oracle and some external optimization algorithm. Pocket2Mol \citep{peng2025pocket2molefficientmolecularsampling} employs a graph neural network, composed of several encoder and predictor modules, that auto-regressively predicts the location and type of each subsequent ligand atom based on existing ligand atoms and the protein pocket. PAFlow \citep{zhou2025priorguidedflowmatchingtargetaware} employs a conditional flow-matching algorithm guided by a learnable number-of-atoms predictor model to generate molecules that better match the size of the binding pocket.

\vspace{ -.7em}
\paragraph{Multi-objective frameworks}
A critical challenge in drug design is multi-objective optimization, as viable drug candidates must satisfy several property criteria simultaneously.
TAGMol \citep{dorna2024tagmoltargetawaregradientguidedmolecule} and DrugDiff \citep{Oestreich2025-vv} use supplementary guide models to influence the Langevin dynamics during sampling, resulting in generated molecules that satisfy multiple property criteria. Graph-GA \citep{Jensen2019} employs an evolutionary algorithm that keeps track of an active population of molecules, applying deterministic crossover and mutation rules to progressively optimize multiple desired properties. OMTRA \citep{dunn2025omtramultitaskgenerativemodel} presents a multi-modal, flexible flow matching model for structure-based drug design. HN-GFN \citep{zhu2023sampleefficientmultiobjectivemolecularoptimization} utilizes a multi-objective Bayesian optimization algorithm combined with a GFlowNet to optimize for several properties, including molecular diversity.

\vspace{ -.7em}
\paragraph{LLMs for molecular generation}
The use of LLMs in drug discovery is currently limited. Current approaches address general-purpose chemistry tasks \citep{bran2023chemcrow, boiko2023autonomous, ma2024ymolmultiscale, choi2026elagenteestructuralartificially}, or fine-tune LLMs to design strong binders in one shot \citep{Sheikholeslami2025}. Genetic algorithms that utilize LLMs for crossover/mutation operations are better suited to the task of property optimization because they are able to incorporate feedback from oracles. MOLLEO \citep{wang2025efficientevolutionarysearchchemical}, which represents the state-of-the-art in LLM-guided drug generation, augments Graph-GA \citep{Jensen2019} by replacing algorithmic crossovers and mutations with LLM-driven structural modifications, and achieves strong results across three protein targets. The authors demonstrate the potential of these inherently chemistry-aware LLMs to be a competitive generative method in drug discovery. A relevant prior work on tool-calling in biochemistry is El Agente Estructural \citep{choi2026elagenteestructuralartificially}, a multimodal framework that equips an LLM with tools for visual and geometric inspection of molecules in 3D space. However, it is designed for human-in-the-loop interaction rather than integration into an optimization loop. 

In this work, we design a minimal yet effective toolbox that improves the quality of LLM-suggested crossover and mutation operations within a genetic algorithm. To our knowledge, this is the first work to utilize LLM tool-calling within an evolutionary framework to optimize molecular properties for drug discovery.

\begin{figure*}[tb]
    \centering
    \includegraphics[width=\linewidth]{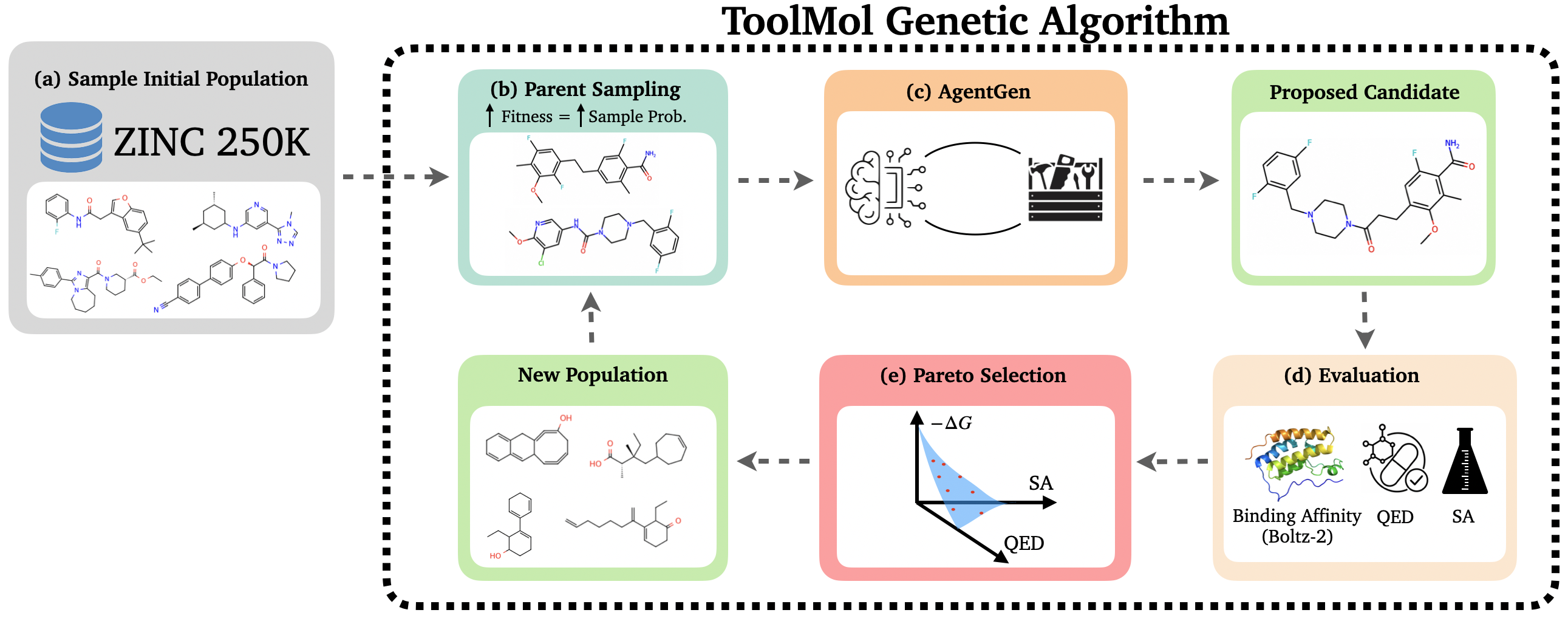}
    \caption{\textbf{Overview of ToolMol.} (a) We sample an initial ligand population from ZINC 250K. (b) Parent ligands are sampled for crossovers \& mutations with probability proportional to their fitness. (c) An agent with access to a set of modification tools generates new ligands using structures from the selected parents. (d) New offspring are evaluated by an oracle for all relevant objectives. (e) A new population is formed from the non-dominated Pareto frontier of the current population. Steps b $\rightarrow$ e are repeated until an oracle budget is reached.} 
    \vspace{ -.7em}
    \label{fig1}
\end{figure*}
\vspace{ -.7em}

\section{ToolMol}
We introduce ToolMol, an agentic, multi-objective genetic algorithm framework that utilizes a tool-calling LLM to make precise and guided modifications on the ligand population. In this section, we first introduce our optimization problem, then describe the genetic algorithm followed by the tool-calling process that comprise ToolMol.

\paragraph{Problem Statement}
We can broadly represent our molecular optimization problem as
$$\M^* =\operatorname*{argmax}_{m \in M} \Phi(m)$$
where $m$ is any valid molecule (ligand) and $M$ is the entire valid chemical space. $\Phi$ is an evaluation function that yields the fitness of $m$ for $n$ objectives. This function can be defined in several ways; one naive approach is to treat the "fitness" of a candidate (the viability or utility of the member) as a weighted sum of all objectives, i.e. $\Phi(m) = \sum_{i} \w_i*\phi_i(m)$, where $\phi_i$ is the $i$th objective evaluator and $\w_i$ is the weight given to that objective.\footnote{Note that if any objective $i$ is minimizing instead of maximizing, we take the negation of the objective evaluator to be $\phi_i$} These weights are arbitrary and can be difficult to choose in practice. Here, we avoid the need to make such arbitrary choices via partial ordering of molecules and the Pareto frontier. Formally, we can compare two molecules by notating $m' \succ m$, meaning that $m'$ strictly dominates $m$ if and only if $\forall i: \phi_i(m') > \phi_i(m)$. We can then define the optimal set $M^*$ to be the non-dominated Pareto frontier, given by $M^* = \{m \in M: \nexists m', m' \succ m\}$, or the set of all molecules that are not dominated by any other molecule. In other words, $M^*$ is constructed by all ligands for which no other ligand strictly exceeds it in every objective. Because the full chemical space $M$ is far too large to ever be sufficiently explored, we search in a limited subspace of $M$ (roughly determined in practice by random initial seeding), and aim to find the non-dominated Pareto frontier $M^*$ within this subspace.

In this work, we consider 3 objectives: the binding affinity ($\Delta G$, in kcal/mol) of $m$ to a particular protein binding target, as estimated with Boltz-2 \citep{Passaro2025}, the quantitative estimate of drug-likeness \citep[QED]{Bickerton2012}, and the estimated synthetic accessibility \citep[SA]{Ertl2009}. All 3 of these objectives are standard targets in generative modeling, with QED and SA ensuring some level of ligand-structure soundness and estimated binding affinity measuring the practical utility of the molecule as a binder of the targeted protein pocket.  
\subsection{Multi-objective Genetic Algorithm}

The underlying framework for the ToolMol algorithm is a multi-objective genetic algorithm (MOGA). The pseudocode for this process is detailed in Algorithm \ref{alg1} (see Figure \ref{fig1} for a visual guide). We begin with an initial population $\mathcal{M}_0$ randomly sampled from the ZINC 250K \citep{Sterling2015} dataset, which provides a good starting base of drug-like \& synthesizable structures. We define an "oracle budget" $B$, which determines how many molecules we evaluate before we terminate the algorithm. We determine the stopping point in this way because predicting binding affinity is generally very computationally expensive; we set a hard limit on the number of total evaluations regardless of the population or offspring size. 

\begin{figure}[H]
\vspace{-15pt}
    \begin{algorithm}[H]
        \caption{ToolMol Genetic Algorithm}
        \label{alg1}
        \SetAlgoLined
        \DontPrintSemicolon
        \KwIn{Initial population $\mathcal{M}_0$, offspring size $n$, oracle budget $B = 1000$}
        \KwOut{All molecule generations $\mathcal{M}_{\text{out}}$}
        
        $\mathcal{M}_c \leftarrow \mathcal{M}_0$ \tcp*{Current population}
        $\mathcal{M}_{\text{out}} \leftarrow \mathcal{M}_0$ \tcp*{All molecules}
        
        \While{\textnormal{oracle budget} $< B$}{
            $\text{offspring} \leftarrow [\,]$\;
        
            \For{$i \leftarrow 1$ \KwTo $n$}{
                Sample $m_0,\, m_1 \sim \mathcal{M}_c$ with probability $\propto k^{\Phi(m)}$ for const $k$\;
                
                $\text{offspring}.\text{append}\bigl(\textcolor{orange}{\hyperref[agentgen]{\textsc{AgentGen}}}(m_0,\, m_1)\bigr)$
            }
            $\mathcal{M}_{\text{out}} \leftarrow \mathcal{M}_{\text{out}} \cup \text{offspring}$\;
            
            $\mathcal{M}_c \leftarrow \mathcal{M}_c \cup \text{offspring}$\;
            
            \For{$m \in \mathcal{M}_c$}{
                Compute $\phi_i(m)$ for each objective $i$\;
            }
        
            $\mathcal{M}_c \leftarrow \textsc{ParetoFrontier}(\mathcal{M}_c)$\;
        }
        \Return $\mathcal{M}_{\text{out}}$\;
    \end{algorithm}
    \vspace{-25pt}
\end{figure}

We sample parent molecules $m_0, m_1$ from the current population with probabilities proportional to $k^{\Phi(m)}$ for some constant $k$. $\Phi(m)$ is the scalar fitness of $m$, given by $\Phi(m) = \sum_if_i(m)$, where $f_i(m)$ is the $i$th objective scaled to $[0, 1]$.\footnote{For the unbounded binding affinity metric, we scale by setting a lower bound of 0 kcal/mol and a safe upper bound of $-13$ kcal/mol, a value that we have never observed our affinity predictor exceed.} We also explore an alternative sampling method based on Pareto ordering, the results of which may be found in Appendix \ref{app:additional_ablations}. We pass the sampled parent molecules into the agentic LLM operator, \textsc{AgentGen}, which is described in the \hyperref[agentgen]{next section}. Specifically, for every pair of sampled parents, \textsc{AgentGen} creates one new candidate, which is added to the current set of offspring. After $n$ new candidates, we merge the current population with the new offspring. We then evaluate all resulting molecules for each objective and form the next generation by taking the non-dominated Pareto frontier of the current population. We continue the evolution until we exhaust our oracle budget.

\subsection{AgentGen: Tool-calling LLM}
\label{agentgen}
Next, we describe the \textsc{AgentGen} function, which represents the tool-calling process that the agentic operator takes to generate new molecules. This process is inspired by the crossover \& mutation operations carried out by classical genetic algorithms, and leverages tool-calling to allow an LLM to act as the sole GA operator.

Given two input molecules $m_1, m_2$, we first format an initial prompt based on our desired objectives. In order to give the LLM sufficient context to execute accurate tool-calls, we append detailed structure information on both input molecules, given by RDKit. This consists of an identification of every atom in each molecule, its RDKit index, number of substitutable hydrogens, neighboring atoms, centrality within the ligand, and more. This information is necessary for the LLM to provide correct parameters for its tool calls, and aids it in making more informed decisions by providing structural context about the input molecules. Full details about the input and intermediate prompt (given by \textsc{PromptFormat}) can be found in Appendix \ref{app:toolmol_prompts}. 

We then begin the tool-calling iteration process, which proceeds for at most \textit{max\_steps} iterations (we use $\textit{max\_steps}=10$ in our experiments, although we note that we rarely ever meet or exceed this threshold). At any given step, the LLM has access to a toolbox of 7 RDKit-backed functions that aid with structural modifications. For all functions, the LLM is responsible for providing all parameters specified in the function definition. If any specification results in an invalid operation (e.g. no available valence), the tool returns a failed state and specifies the particular error.
\begin{center}
   \includegraphics[width=\linewidth]{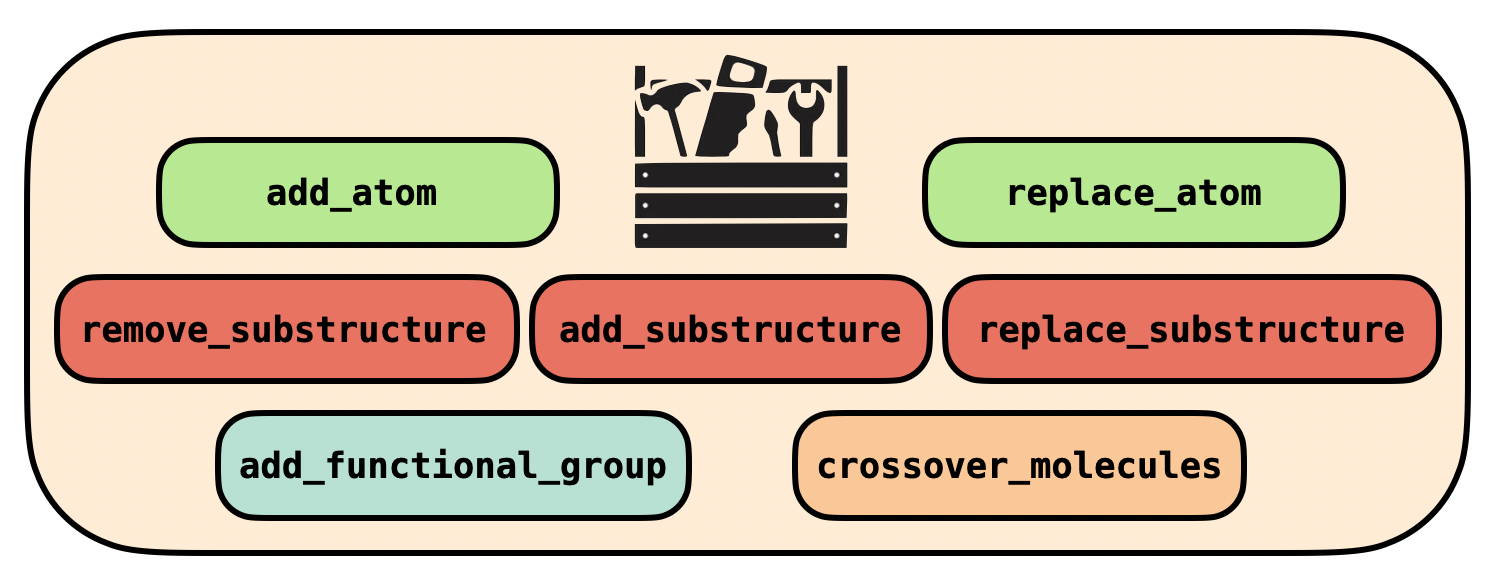} 
\end{center}
\vspace{-10 pt}
 The LLM-callable functions are listed in the graphic above. For complete details on each tool and its function parameters, see Appendix \ref{app: toolmol_toolbox}.

The LLM is encouraged to call \texttt{crossover\_molecules} on the first step, when initially passed two input molecules from the parent population. On subsequent steps, the LLM is encouraged to use other tools on the molecule resulting from the crossover operation. This simulates how a standard genetic algorithm typically performs a crossover on two parent candidates, then an optional mutation on the resulting offspring \citep{Jensen2019}.

\begin{figure*}[t]
    \centering
    \includegraphics[width=\linewidth]{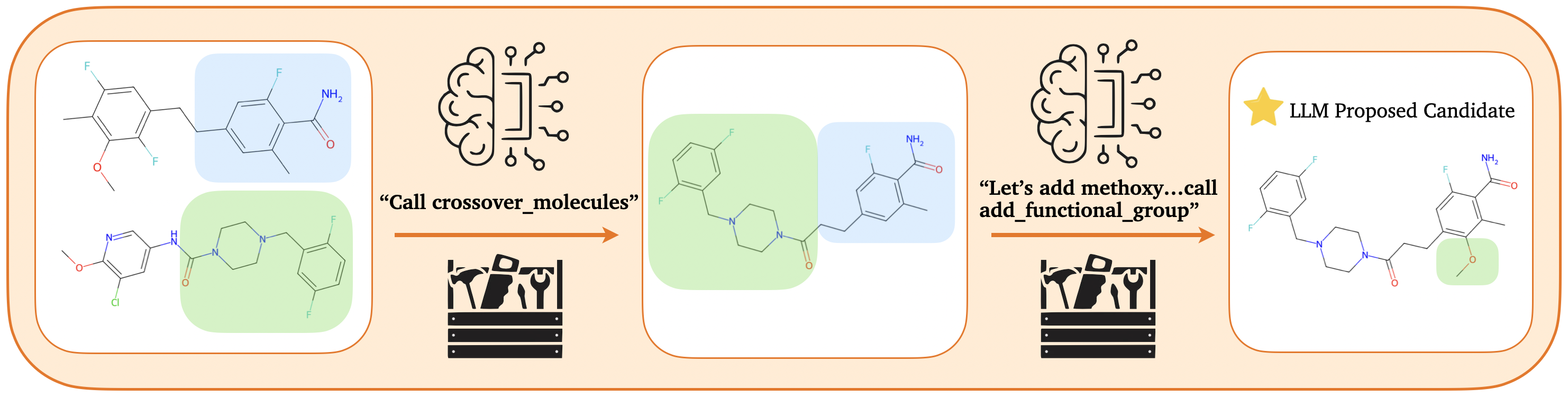}
    \caption{\textbf{An example tool-calling process.} The agent first decides to perform a crossover on the input molecules, utilizing \texttt{crossover\_molecules}. Then it decides to attach a methoxy group to the benzene structure, utilizing \texttt{add\_functional\_group}. At this point, it decides that the modifications are sufficient, and the new molecule is added to the offspring population.} 
    \label{fig2}
    
\end{figure*}   

A single tool call either succeeds and returns the new modified molecule, or fails and returns a message detailing the reason for the error. In both cases, information about the executed tool call and structural details about the new molecule are added to the conversation history for the next tool-calling iteration. This process repeats until we either hit the \textit{max\_steps} iteration budget or until the LLM decides that it has made sufficient modifications. Because all modifications are made in the deterministic graph space defined by the RDKit \texttt{Mol} object, the final molecule returned by this process is guaranteed to be a valid molecule with a valid SMILES encoding. The only exception is if the LLM fails to call any function correctly for \textit{max\_steps} iterations, which we observe to be extremely uncommon. A short example of this process is demonstrated in Figure \ref{fig2}.

\section{Experiments}
\subsection{Experimental Setup}

We evaluate the effectiveness of ToolMol on the multi-objective task of optimizing for protein-ligand binding affinity while preserving drug-likeness and synthetic accessibility.
\vspace{ -.7 em}
\paragraph{Targets.}
In this work, we focus on three functionally \& structurally unique protein-binding targets:
\begin{enumerate}
    \item \textbf{c-MET} (MET\_HUMAN): Hepatocyte growth factor receptor
    \item \textbf{BRD4} (BRD4\_HUMAN): Bromodomain-containing protein 4
    \item \textbf{ACAA1} (THIK\_HUMAN): 3-ketoacyl-CoA thiolase, peroxisomal
\end{enumerate}

The targets c-MET and BRD4 have significant medicinal chemistry literature \citep{Hong2016, Organ2011}, while ACAA1 has not been significantly explored as a drug target. In particular, ACAA1 has no associated experimental binding-affinity measurements in BindingDB \citep{Liu2007}, a database of experimentally measured interactions between drug-target proteins and ligands. Thus, the results for ACAA1 report on the performance of the LLM on a target that has not appeared frequently within its pretraining dataset.
\vspace{ -.7 em}
\paragraph{Pipeline.}
For ToolMol, we seed 60 small molecules from ZINC 250K \citep{Sterling2015} to comprise our initial population, with an offspring size of 35. We utilize an exponential constant $k=10$ for the parent sampling step. For all LLM-based components, including baselines, we use GPT-OSS-120B \citep{openai2025gptoss120bgptoss20bmodel}. We estimate ligand-protein binding affinities with the recent biomolecular foundation model Boltz-2 \citep{Passaro2025}. This decision is primarily motivated by the high accuracy Boltz-2 demonstrates in favoring molecules that score highly on gold-standard Absolute Binding Free Energy \citep[ABFE]{Feng2022} metrics. We provide a brief correlation analysis between predicted Boltz-2,
ABFE, and AutoDock \citep{Trott2009} scores in Appendix \ref{correlation_analysis} to further justify this decision. 

\vspace{ -.7 em}
\paragraph{Baselines.}
We evaluate ToolMol against the following methods. 
\begin{enumerate}
    \item \textbf{Pocket2Mol}~\citep{peng2025pocket2molefficientmolecularsampling} E(3)-equivariant autoregressive model that generates 3D molecules conditioned on a protein pocket via diffusion.
    \item \textbf{TAGMol} \citep{dorna2024tagmoltargetawaregradientguidedmolecule} 3D structure-based framework that decouples diffusion sampling from gradient-based property guidance, using predicted binding affinity, QED, and SA to steer generation. 
    \item \textbf{PAFlow}~\citep{zhou2025priorguidedflowmatchingtargetaware} Conditional flow matching method that leverages a protein-ligand interaction predictor to guide generation toward high-affinity, drug-like molecules.
    \item  \textbf{Graph-GA} \citep{Jensen2019} Genetic algorithm operating via predefined crossover and mutation rules on molecular graphs. 
    \item \textbf{ShinkaEvolve} \citep{lange2025shinkaevolveopenendedsampleefficientprogram}: We adapt the hybrid MAP-Elites \citep{mouret2015illuminatingsearchspacesmapping} and islands algorithm from ShinkaEvolve, an LLM-based evolutionary approach which achieved fantastic results in algorithm design. We adapt this method for drug design, and test variants with both MOLLEO-style LLM mutations and the ToolMol toolbox (details of our implementation are in Appendix~\ref{app: shinkaevolve}).
    \item \textbf{MOLLEO} \citep{wang2025efficientevolutionarysearchchemical} MOLLEO extends Graph-GA by replacing predefined genetic operators with an LLM that directly performs crossovers and mutations on molecular candidates, optimizing jointly over binding affinity, QED, and SA.
    
\end{enumerate}

We note that Pocket2Mol, TAGMol, and PAFlow are designed for oracle-free inference sampling, while all other baseline methods explicitly use affinity feedback as guidance during search. We also note that we use Boltz-2 for affinity prediction for all relevant baselines.

\begin{table*}[t!]
\centering
\caption{Results of ToolMol compared to generative modeling and LLM-based baselines across three protein targets: c-MET, BRD4, and ACAA1. N/A indicates zero observations across all seeded runs. Best results are \textbf{bolded}, second best are \underline{underlined}}
\label{main_results}
\small
\resizebox{\linewidth}{!}{%
\begin{tabular}{l|cccccccc}
\toprule
Metric & Pocket2Mol & TAGMol & PAFlow & Graph-GA & \makecell{ShinkaEvolve \\ (No Tools)} & \makecell{ShinkaEvolve \\ (Tools)} & MOLLEO & \makecell{ToolMol \\(ours)} \\
\midrule
\multicolumn{9}{c}{\textbf{c-MET} (MET\_HUMAN) } \\
\midrule
BA ($\downarrow$)  & $\underline{-11.27 \pm 0.29}$ & $\mathbf{-11.45 \pm 0.11}$ & $-10.63 \pm 0.02$ & $-10.21 \pm 0.26$ & $-10.17 \pm 0.28$ & $-11.08 \pm 0.07$ & $-10.15 \pm 0.19$ & $-11.00 \pm 0.09$ \\
FA ($\downarrow$) & $-9.39 \pm 0.27$ & $-7.39 \pm 0.43$ & $-7.58 \pm 0.44$ & $-9.19 \pm 0.29$ & $\underline{-9.72 \pm 0.14}$ & $\underline{-9.72 \pm 0.35}$ & $-9.62 \pm 0.11$ & $\mathbf{-10.35 \pm 0.17}$ \\
HV ($\uparrow$)     & $0.58 \pm 0.007$ & $0.56 \pm 0.005$ & $0.50 \pm 0.0006$ & $0.57 \pm 0.01$ & $0.58 \pm 0.01$ & $0.59 \pm 0.008$ & $\underline{0.60 \pm 0.01}$ & $\mathbf{0.62 \pm 0.01}$ \\
\midrule
\multicolumn{9}{c}{\textbf{BRD4} (BRD4\_HUMAN) } \\
\midrule
BA ($\downarrow$) & $-10.02 \pm 0.24$ & $-9.54 \pm 0.10$ & $-8.33 \pm 0.09$ & $-9.79 \pm 0.26$ & $-9.59 \pm 0.09$ & $\mathbf{-10.80 \pm 0.19}$ & $-9.87 \pm 0.23$ & $\underline{-10.64 \pm 0.28}$ \\
FA ($\downarrow$) & $-8.61 \pm 0.17$ & $-8.06 \pm 0.08$ & $-7.54 \pm 0.12$ & $-9.07 \pm 0.31$& $-9.38 \pm 0.03$ & $-9.20 \pm 0.07$ & $\underline{-9.48 \pm 0.19}$ & $\mathbf{-9.91 \pm 0.18}$ \\
HV ($\uparrow$)       & $0.52 \pm 0.02$ & $0.53 \pm 0.008$ & $0.43 \pm 0.01$ & $0.56 \pm 0.02$ & $0.56 \pm 0.008$ & $0.57 \pm 0.001$ & $\underline{0.59 \pm 0.01}$ & $\mathbf{0.60 \pm 0.01}$ \\
\midrule
\multicolumn{9}{c}{\textbf{ACAA1} (THIK\_HUMAN) } \\
\midrule
BA ($\downarrow$)  & $-8.45 \pm 0.53$ & $-8.58 \pm 0.06$ & $-7.90 \pm 0.02$ & $-8.81 \pm 0.13$ & $-8.78 \pm 0.31$ & $\mathbf{-10.20 \pm 0.11}$ & $-8.41 \pm 0.41$ & $\underline{-9.70 \pm 0.23}$ \\
FA ($\downarrow$) & $-7.39 \pm 0.37$ & $-6.67 \pm 0.09$ & N/A & $-8.05 \pm 0.16$ & $\underline{-8.51 \pm 0.19}$ & $-8.11 \pm 0.08$ & $-8.12 \pm 0.45$ & $\mathbf{-8.78 \pm 0.15}$ \\
HV ($\uparrow$)  & $0.48 \pm 0.09$ & $0.46 \pm 0.004$ & $0.33 \pm 0.02$ & $0.50 \pm 0.01$ & $0.51 \pm 0.005$ & $\underline{0.53 \pm 0.01}$ & $0.51 \pm 0.02$ & $\mathbf{0.54 \pm 0.008}$ \\
\midrule
Avg. Rank ($\downarrow$) & $5.00$ & $6.11$ & $7.56$ & $5.00$ & $3.89$ & $\underline{2.56}$ & $3.89$ & $\textbf{1.56}$ \\
\bottomrule
\end{tabular}
}
\end{table*}

\vspace{ -.7 em}
\paragraph{Evaluation Metrics.}
We consider the following evaluation criteria:
\begin{itemize}

    \item \textbf{Binding Affinity (BA)}: Mean binding affinity (kcal/mol) of the top 10 strongest binding molecules to the particular protein target, predicted by Boltz-2.
    \item \textbf{Filtered Affinity (FA)}: We further filter our ligand sample by only considering ligands that satisfy sufficient Quantitative Estimate of Drug-likeness \citep[QED]{Bickerton2012} and Synthetic Accessibility \citep[SA]{Ertl2009} scores. We filter by $\text{QED} > 0.5$ and $\text{SA} < 3.0$, then take the mean binding affinity of the top 10 strongest binding molecules that survive this filter. This is a crucial metric that measures the strength of generated ligands that may actually pass the first stage of a real-world wet lab synthesis.
    \item  \textbf{Hypervolume (HV)}: Measures the Euclidean volume of the 3D-space (affinity, QED, SA) covered by the non-dominated Pareto frontier formed by the set of all generated molecules. We scale all objectives to $[0, 1]$ and use a reference point of $(1, 1, 1)$ for our calculations. 
\end{itemize}

\vspace{ -.7 em}
\paragraph{Results.}
Table \ref{main_results} shows the results of running all baselines and ToolMol on all 3 protein targets, along with a Quantitative Estimate of Drug-likeness \citep[QED]{Bickerton2012} maximization objective and a Synthetic Accessibility \citep[SA]{Ertl2009} minimization objective. We aim to generate molecules that yield the strongest possible binding affinity, that also simultaneously maintain strong-enough QED and SA properties. This is motivated by processes in real-world drug discovery pipelines, where molecules are strongly optimized for binding affinity, but must also be sufficiently drug-like and synthesizable to be realistic candidates. We run all GA-based methods (Graph-GA, MOLLEO, ToolMol) on 5 different seeded sets of initial molecules, and ShinkaEvolve on 3 different seeded sets. Both methods terminate after 1000 Boltz-2 oracle evaluations. We generate 3 seeds of 1000 sampled molecules from Pocket2Mol, TAGMol, and PAFlow for each target, matching the oracle budget for the GA \& ShinkaEvolve.

All metrics are reported on a sample of the entire generated ligand pool. For each method, we first Butina cluster the full pool of all generated molecules (with similarity threshold = 0.6), then from each resulting cluster, we take the molecule with the strongest binding affinity in that cluster. This way, we most effectively assess the quality of all structurally unique generations, encouraging diversity in results and favoring consistent strong metrics across a wide region of chemical space.

ToolMol achieves the best average rank across seven methods and nine metrics. Notably, it outperforms every baseline in both multi-objective metrics: filtered mean and hypervolume. We consider these metrics to be the most important, as they are most pertinent to our multi-objective problem statement. ToolMol also consistently outperforms MOLLEO in single-objective binding affinity. Additionally, integrating the ToolMol toolbox into ShinkaEvolve yields the strongest binding affinity scores on two targets. This demonstrates the generalizability of our tool-calling framework, as our toolbox yields consistent improvements when integrated into two distinct optimization algorithms: classical genetic algorithms and MAP-Elites.

We note that while certain generative modeling baselines such as Pocket2Mol and TAGMol exceed our method in pure single-objective affinity, the drastic drop in filtered binding affinity for those methods reveals that the crucial QED and SA properties are not sufficiently fulfilled. This implies that the majority of the high scoring affinity compounds are not drug-like or synthesizable enough to be practical. This is further supported by the low hypervolume scores for these generative baselines. Out of all tested methods, ToolMol is the most successful at creating molecular candidates that balance high binding affinity with desirable secondary objectives, reflecting high real-world utility as a generative framework.

\begin{table*}[t!]
\centering
\caption{ABFE results of top 15 molecules for ToolMol and MF-LAL \citep{eckmann2025mflaldrugcompoundgeneration}. ToolMol achieves significantly higher ABFE scores for both sets of evaluated molecules.}
\begin{adjustbox}{width=1\textwidth}
\setlength{\tabcolsep}{4pt}
\begin{tabular}{l|ccc|ccc}
\toprule
\multirow{2}{*}{Method}
& \multicolumn{3}{c|}{c-MET}
& \multicolumn{3}{c}{BRD4} \\
\cmidrule(lr){2-4} \cmidrule(lr){5-7}
 & ABFE ($\downarrow$) & QED ($\uparrow$) & SA ($\downarrow$) & ABFE ($\downarrow$) & QED ($\uparrow$) & SA ($\downarrow$) \\
\midrule
MF-LAL & $-6.7 \pm 3.1$ & $\underline{0.63 \pm 0.15}$& $3.50 \pm 0.58$ & $-6.2 \pm 3.9$ & $\underline{0.59 \pm 0.07}$ & $3.60 \pm 0.55$ \\
ToolMol & $\mathbf{-7.96 \pm 2.77}$ & $0.45 \pm 0.21$ & $\underline{3.26 \pm 0.36}$ & $\mathbf{-8.4 \pm 3.9}$ & $0.27 \pm 0.18$ & $\underline{3.37 \pm 0.45}$ \\
ToolMol (filtered) & $\underline{-7.3 \pm 3.8}$ & $\mathbf{0.66 \pm 0.08}$ & $\mathbf{2.75 \pm 0.16}$ & $\underline{-6.4 \pm 3.5}$ & $\mathbf{0.63 \pm 0.10}$ & $\mathbf{2.80 \pm 0.14}$ \\
\bottomrule
\end{tabular}
\end{adjustbox}
\label{abfe}
\end{table*}

\subsection{Absolute Binding Free Energy (ABFE)}
To further demonstrate the usefulness of ToolMol for real-world drug design, we compute Absolute Binding Free Energy (ABFE) scores for its generated molecules. ABFE uses expensive molecular dynamics simulations to accurately calculate binding free energy \citep{Heinzelmann2021}. It is the current gold-standard for computational binding affinity prediction, and thus reflects a higher degree of accuracy in predicting real-world experimental activity. We benchmark against MF-LAL \citep{eckmann2025mflaldrugcompoundgeneration}, a state-of-the-art multi-fidelity approach to drug design that specifically targets ABFE scores through high-fidelity guided VAE decoding. Following the exact ABFE setup from MF-LAL, we evaluate ToolMol on the two targets reported in the MF-LAL paper, c-MET and BRD4. We use two sets of top 15 molecules from ToolMol, one ranked solely on binding energy and the other after applying the QED $> 0.5$, SA $< 3.0$ filter. These results are shown in Table \ref{abfe}. Additional details about parameters used in these ABFE calculations can be found in Appendix \ref{app:abfe_setup}.

The top 15 molecules ranked by Boltz-2 predicted affinity achieve strong ABFE scores for both targets, surpassing MF-LAL by a large margin. This comes at a modest cost to secondary objectives, though these molecules still exceed MF-LAL in synthesizability. The top 15 filtered ligands yield slightly weaker ABFE scores, but still beat MF-LAL in every metric, scoring higher on ABFE while maintaining more desirable QED and SA values. Notably, although ABFE feedback is not explicitly included within our optimization pipeline, we outperform the current state-of-the-art method for high ABFE-scoring molecules simply by optimizing Boltz-2 predicted affinity with a tool-assisted LLM. The fact that LLM-generated ligands can achieve state-of-the-art results in this area demonstrates great potential for LLMs to have a real, significant impact in computational drug discovery.

\subsection{Ablations}
\label{main_ablations}

\begin{table*}[t]
\centering
\caption{Ablations: MOLLEO's invalid generation rate and ToolMol's genetic algorithm. Isolating the impact of the ToolMol toolbox reveals that the tool-calling process significantly improves results.}
\label{ablation_merged}
\resizebox{\textwidth}{!}{%
\begin{tabular}{ll|cccc}
\toprule
Target & Metric & MOLLEO & \makecell{MOLLEO \\ (Retry Failures)} & \makecell{ToolMol \\ (MOLLEO GA)} & ToolMol \\
\midrule
\multirow{3}{*}{c-MET}
 & Binding Affinity ($\downarrow$)  & $-10.15 \pm 0.19$ & $-9.98 \pm 0.31$  & $\mathbf{-11.14 \pm 0.20}$ & $\underline{-11.00 \pm 0.09}$ \\
 & Filtered Affinity ($\downarrow$) & $-9.62 \pm 0.11$  & $-9.72 \pm 0.24$ & $\underline{-10.22 \pm 0.16}$ & $\mathbf{-10.35 \pm 0.17}$ \\
 & Hypervolume ($\uparrow$)         & $\underline{0.60 \pm 0.01}$ & $0.57 \pm 0.01$ & $\underline{0.60 \pm 0.02}$ & $\mathbf{0.62 \pm 0.01}$ \\
\midrule
\multirow{3}{*}{BRD4}
 & Binding Affinity ($\downarrow$)  & $-9.87 \pm 0.23$  & $-9.67 \pm 0.31$  & $\underline{-10.61 \pm 0.33}$ & $\mathbf{-10.64 \pm 0.28}$ \\
 & Filtered Affinity ($\downarrow$) & $-9.48 \pm 0.19$  & $-9.43 \pm 0.26$  & $\underline{-9.87 \pm 0.32}$  & $\mathbf{-9.91 \pm 0.18}$ \\
 & Hypervolume ($\uparrow$)         & $\underline{0.59 \pm 0.01}$ & $0.56 \pm 0.02$ & $\underline{0.59 \pm 0.01}$ & $\mathbf{0.60 \pm 0.01}$ \\
\midrule
\multirow{3}{*}{ACAA1}
 & Binding Affinity ($\downarrow$)  & $-8.41 \pm 0.41$  & $-8.04 \pm 0.12$  & $\mathbf{-9.87 \pm 0.18}$  & $\underline{-9.70 \pm 0.23}$ \\
 & Filtered Affinity ($\downarrow$) & $-8.12 \pm 0.45$  & $-7.93 \pm 0.11$  & $\underline{-8.77 \pm 0.24}$ & $\mathbf{-8.78 \pm 0.15}$ \\
 & Hypervolume ($\uparrow$)         & $0.51 \pm 0.02$   & $0.49 \pm 0.01$   & $\underline{0.53 \pm 0.02}$ & $\mathbf{0.54 \pm 0.008}$ \\
\midrule
 & Avg. Rank ($\downarrow$)      & $2.89$ & $3.87$ & $\underline{1.78}$ & $\mathbf{1.22}$ \\
\bottomrule
\end{tabular}
}
\end{table*}

We present ablations that specifically highlight the impact of the tool-calling process, demonstrating the isolated impact of the toolbox provided to the LLM in ToolMol. We compare with MOLLEO, which is the closest methodological neighbor to ToolMol.

First, we ablate the effect of the underlying genetic algorithm on the results. We run ToolMol on the exact MOLLEO genetic algorithm (MOLLEO GA); this isolates the particular impact of introducing function-calling to the crossover/mutation process by removing algorithmic differences of the underlying framework. Second, we ablate the consequences of the high invalid molecule generation rate faced by MOLLEO. We observe that across 1000 molecule generations, MOLLEO will consistently yield $\sim$350 invalid generations, due to formatting issues or syntactically invalid SMILES. For each invalid generation, MOLLEO immediately falls back on the default Graph-GA crossover / mutation crossovers. We can naively reduce the MOLLEO failed generation rate simply by forcing the LLM to retry its generation until it yields a valid result. This gives a stronger comparison between the MOLLEO generation process and the ToolMol function-calling process per 1000 generations by eliminating a large portion of the extraneous Graph-GA impact within MOLLEO. We give the LLM a maximum of 10 retry steps.

Table \ref{ablation_merged} compares the original MOLLEO \& ToolMol with the two aforementioned ablations. We discuss two interesting results. First, simply integrating ToolMol's toolbox into MOLLEO's genetic algorithm alone yields significant improvements in binding affinity and QED/SA over MOLLEO's LLM-based modifications. Second, we report that the retry method for MOLLEO drops invalid LLM generations down to nearly $0\%$, yielding a single digit number of invalid strings every 1000 generations. This means that the majority of the final ligand pool is generated by the MOLLEO LLM operator. However, we observe that this does not improve the performance on our evaluation targets, and in fact degrades performance across nearly every metric. Thus, we further isolate the effect of the ToolMol function-calling process by focusing solely on the LLM operator in MOLLEO, and demonstrate that ToolMol's agent-generated ligands are still superior to MOLLEO's LLM-generated ligands in all metrics.

\begin{figure*}[b]
    \centering
    \caption{ToolMol \& MOLLEO modification steps and reasoning traces. MOLLEO fails to execute its planned modifications, while ToolMol successfully executes its ideas.}
    \includegraphics[width=\linewidth]{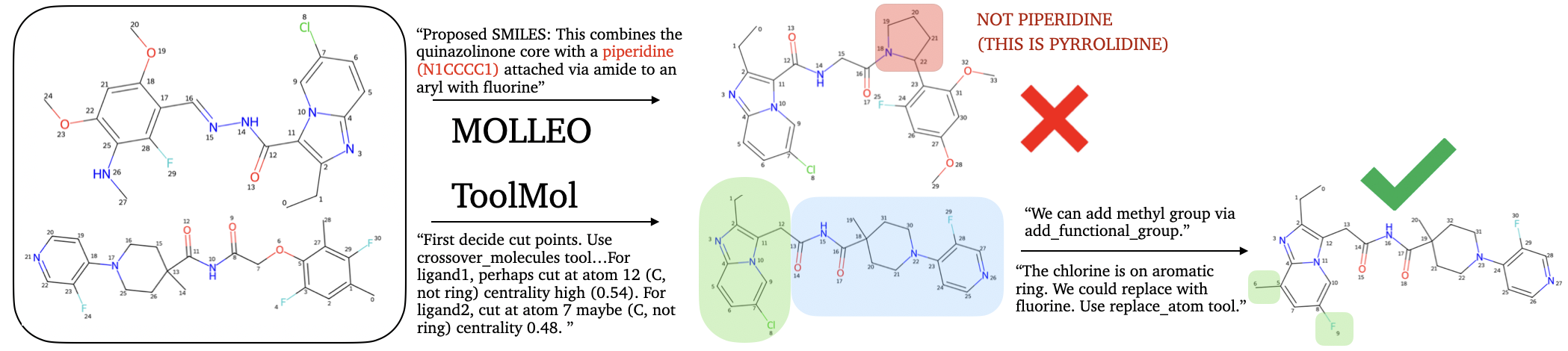}
    \label{case_study_fig}
\end{figure*}

\subsection{Why does tool-calling improve performance?}
To understand why tool-calling benefits this black-box optimization problem, we simulate a single LLM modification step by providing 2 fixed input molecules (sampled from ZINC 250K), and compare the reasoning traces of GPT-OSS-120B using the ToolMol toolbox to GPT-OSS-120B using the MOLLEO modification scheme. Figure \ref{case_study_fig} shows how the input molecules are modified by these two methods and examines the modifications against the corresponding reasoning traces.

We observe that in the MOLLEO process, there are critical discrepancies between the planned modifications described in the LLM's reasoning trace and the actual resulting molecule. In contrast, every modification made to the molecule by ToolMol is exactly consistent with what the LLM describes in its reasoning trace. Full reasoning traces for this case study can be found in Appendix \ref{app:case_study}.

For further quantitative confirmation, we repeat this experiment on 10 more pairs of distinct input ligands. Out of 10 generations, ToolMol yields two processes where there is a some discrepancy between the reasoning trace and the resulting modification, while MOLLEO yields seven such erroneous processes, a significant difference ($p=0.02$, by 2-sided independent t-test). The two ToolMol errors arise from slightly imprecise parameters passed into the tools (e.g. calling \texttt{replace\_atom} with the same element that already exists at that index) leading to a non-matching change, while MOLLEO frequently misidentifies structures and inserts incorrect groups into the output. Thus, in general, the resulting compound modifications generated with ToolMol better match the desired changes outlined by the LLM in its reasoning trace. We conjecture that, by reducing the potential for error between LLM reasoning and the generated compounds, ToolMol takes better advantage of the vast chemical knowledge that LLMs naturally possess through pretraining. We believe that this is largely why ToolMol achieves significantly stronger binding affinity results on every protein target that we tested on.

\section{Discussion \& Conclusion}
We present ToolMol, an agentic multi-objective drug discovery framework that iteratively optimizes small molecule ligands for protein binding. We build a Pareto-optimizing genetic algorithm that utilizes an exponential-sampling procedure, and combine it with an LLM that has access to a structured toolbox of seven deterministic, RDKit-backed operations. Rather than requiring an LLM to directly generate or modify molecular string encodings (a task prone to syntactic failure), this agentic framework reduces the potential failure surface by abstracting away the necessity for the LLM to be syntactically perfect in its outputs. We achieve state-of-the-art results in three protein-ligand binding tasks, consistently generating molecules that outscore baselines in predicted binding affinity, QED, and SA. Despite not being directly optimized for ABFE score anywhere in its pipeline, ToolMol also achieves exceptional results in this area, markedly increasing the power of LLMs as tools for computational drug discovery. We hypothesize that the inherent chemical knowledge that LLMs hold benefits them in designing more realistic molecules, perhaps more similar to what an actual medicinal chemist might synthesize. This gives them a distinct advantage over recent generative models, which often generate compounds that lack desired molecular properties, demonstrated by the poor multi-objective metric scores achieved by diffusion and flow-based methods on our task.

\vspace{-10pt}
\paragraph{Limitations} It is important to note the existence of concerns about the accuracy of Boltz-2 as an affinity predictor. In particular, recent studies have shown that the performance of Boltz-2 degrades significantly when evaluating on novel, out-of-distribution ligand scaffolds and protein targets \citep{Li2026-ec, Shepard2026}. Nonetheless, there is evidence that Boltz-2 outperforms the primary industry alternative, AutoDock Vina, in pose and affinity prediction \citep{yue2026comparative}. We also have observed that Boltz-2 shows better agreement with well-regarded Absolute Binding Free Energy calculations than AutoDock, evaluated on one of our own relevant protein targets (see Appendix \ref{correlation_analysis}).

\vspace{-3mm}
\section*{Impact Statement}
This work has the potential to accelerate the early stages of drug discovery by enabling more efficient identification of high-affinity, drug-like ligand candidates. Impacts include reducing the time and cost of lead optimization, improving the quality of computationally generated drug candidates entering wet-lab validation, and providing a modular, interpretable framework where an LLM's reasoning for each molecular modification is transparent and traceable through tool calls. 
\section*{Author Disclosures}
M.K.G. has an equity interest in and is a cofounder and scientific advisor of VeraChem LLC. He is also on the scientific advisory boards of Denovicon Therapeutics, In Cerebro, Cold Start Therapeutics, and Beren Therapeutics. R.Y and P.E have equity interests and are co-founders of Aethermol LLC.
\vspace{-3mm}
\section*{Acknowledgment}
This work was  supported in part by the U.S. Army Research Office
under Army-ECASE award W911NF-07-R-0003-03, the U.S. Department Of Energy, Office of Science, IARPA HAYSTAC Program, and NSF Grants \#2205093, \#2146343, \#2134274, \#2441832, CDC-RFA-FT-23-0069, DARPA AIE FoundSci and DARPA YFA. 

\bibliography{references}
\bibliographystyle{icml2026}

\newpage
\appendix
\onecolumn

\section{Case Study}
\label{app:case_study}
We provide further details on the case study presented in Figure \ref{case_study_fig} of the main paper. Below are the initial two molecules given to both MOLLEO and ToolMol.

\begin{figure}[h]
    \centering
    \includegraphics[width=0.48\textwidth]{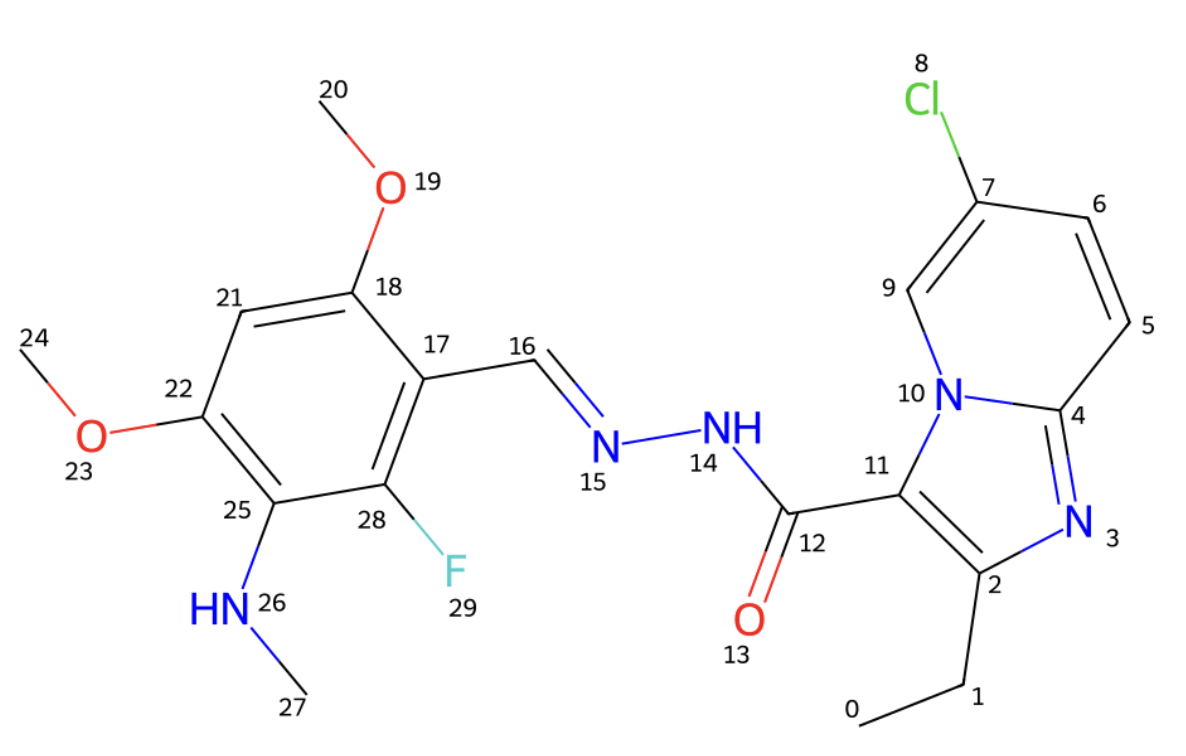}
    \hfill
    \includegraphics[width=0.48\textwidth]{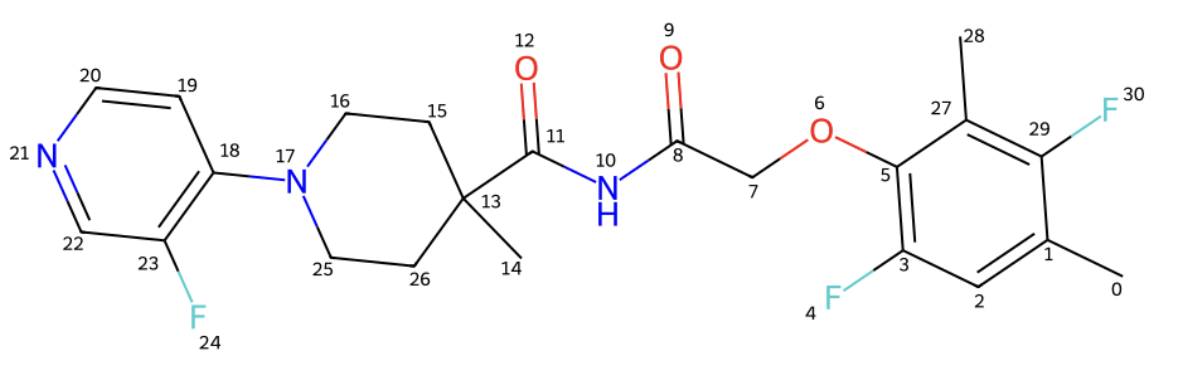}
\end{figure}

First, we see how the LLM makes modifications \textbf{without} the ToolMol toolbox. Here are excerpts from its reasoning trace that describe how it intends to modify the molecule:
\begin{tcolorbox}[
  colback=brown!8!white,
  colframe=brown!45!white,
  coltitle=brown!60!black,
  colbacktitle=brown!20!white,
  boxrule=0.5pt,
  arc=2pt,
  left=4pt,
  right=4pt,
  top=4pt,
  bottom=4pt,
title=MOLLEO Reasoning Trace
]
"...We can propose a hybrid: combine quinazoline core with a morpholine or piperazine side chain (like in many kinase inhibitors). Add halogen for potency, add -OCH3 groups for interactions. Use a hinge-binding motif: a N,N-dimethylcarboxamide? Actually c-MET hinge binder often uses 4-anilino quinazoline. We could take quinazoline from first (c1nc2ccc(Cl)cn2c1C=O), but modify the N-linked hydrazone to an anilino group with fluorine. Add a piperidine amide from second. Proposed SMILES: something like: "CCc1nc2ccc(Cl)cn2c1C(=O)NCC(=O)N1CCCC1c1ccc(F)cc1O". This combines the quinazolinone core with a piperidine (N1CCCC1) attached via amide to an aryl with fluorine...Let\'s design stepwise: Core: CCc1nc2ccc(Cl)cn2c1C(=O)N — attach a linker: N-CH2-CH2-C(=O)-N1CCCC1 (piperidine). Then attach a phenyl ring to piperidine nitrogen...Thus SMILES: "CCc1nc2ccc(Cl)cn2c1C(=O)NCC(=O)N1CCCC1c1c(F)cc(OC)cc1OC"."
\end{tcolorbox}

Below is a visualization of the molecule generated from this process:
\begin{figure}[h]
    \centering
    \caption{This figure shows the initial molecules, and resultant molecules after LLM modifications using MOLLEO and ToolMol. We see that MOLLEO fails to generate the required molecule}
    \includegraphics[width=0.48\textwidth]{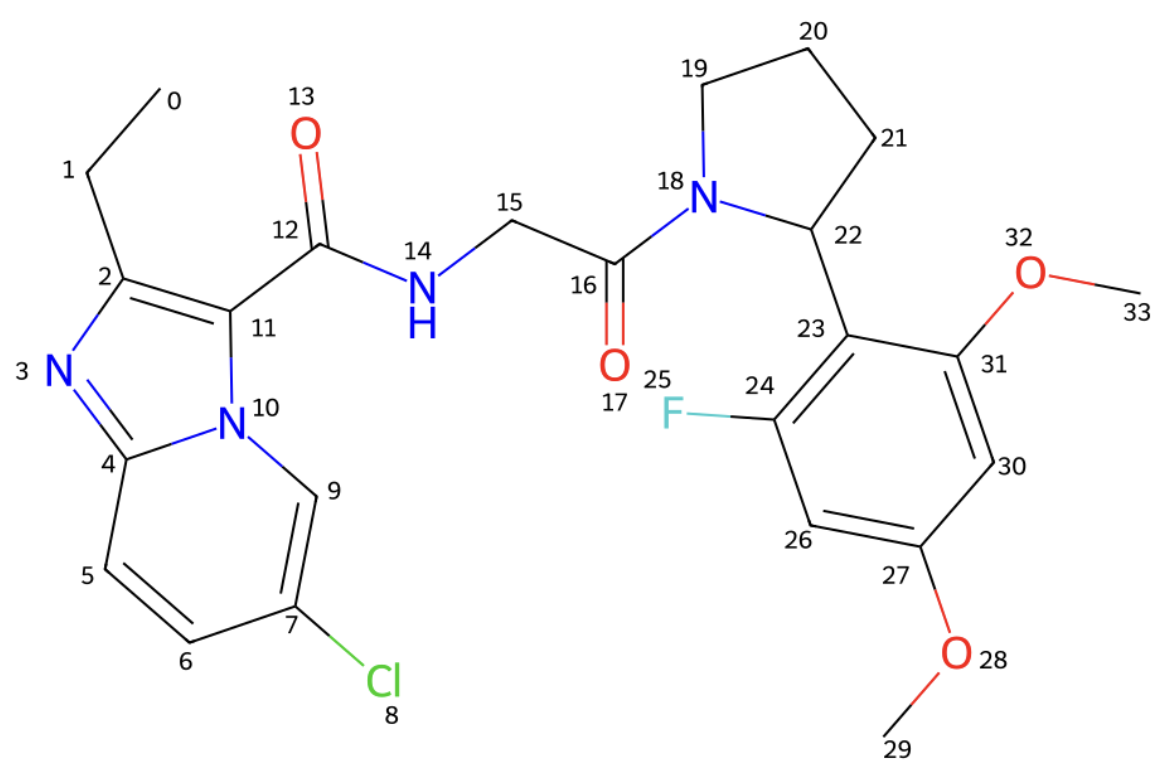}
\end{figure}

We observe that while many parts of the final molecule are consistent with what is described by the reasoning trace, there are certain parts that are entirely inconsistent with the LLM's planned modifications. For instance, it insists that it is using the piperidine ring from ligand 2, but the ring at indices 18-22 is actually a 5-membered pyrrolidine ring; it hallucinates the SMILES for piperidine multiple times throughout its reasoning trace. The resulting molecule is actually a reasonable crossover between the inputs, but is simply not fully consistent with the reasoning trace that generated it. We observe this trend commonly within MOLLEO generations, where the resulting modification(s) do not exactly align with the desired theoretical modifications that the LLM envisioned.

We contrast this to the ToolMol function-calling process. Here are the reasoning traces for multiple tool steps.

\begin{tcolorbox}[
    breakable,
  colback=brown!8!white,
  colframe=brown!45!white,
  coltitle=brown!60!black,
  colbacktitle=brown!20!white,
  boxrule=0.5pt,
  arc=2pt,
  left=4pt,
  right=4pt,
  top=4pt,
  bottom=4pt,
title=ToolMol Reasoning Trace
]
Need to use tools to make modifications. First decide cut points. Use crossover\_molecules tool. Choose roughly half indices with high centrality not in ring? Must not split at ring index...For ligand1, perhaps cut at atom 12 (C, not ring) centrality high (0.54). For ligand2, cut at atom 7 maybe (C, not ring) centrality 0.48. Use those. We\'ll call crossover.\\

...Typically, we might want to increase H-bond donors/acceptors, maybe add a fluorine, or replace a chlorine with fluorine to improve metabolic stability. Also maintain drug-like properties...Look at structure: There\'s a fused heterocycle (maybe quinazoline). The chlorine is on aromatic ring. We could replace with fluorine. Use replace\_atom tool.\\

...Goal: better c-MET binding maybe increase H-bond donors/acceptors, aromaticity, add fluorine, etc. Looking at structure: It has fused heteroaryl (pyrimidine) etc. Maybe add a methyl group on the phenyl ring to increase hydrophobic interaction. Which atom index corresponds to aromatic carbon with substitution? Let\'s inspect aromatic ring indices...So we can add a methyl to a carbon with a hydrogen, e.g., atom 5 has substitutable\_hydrogens 1 (per data). It\'s a carbon in aromatic ring with one H. So we can add methyl group via add\_functional\_group.
\end{tcolorbox}
This describes a 3-step function-calling process that the LLM carried out, first executing a crossover, then making two minor modifications to the resulting molecules. Below is the final generated molecule:
\begin{figure}[h]
    \centering
    \includegraphics[width=0.48\textwidth]{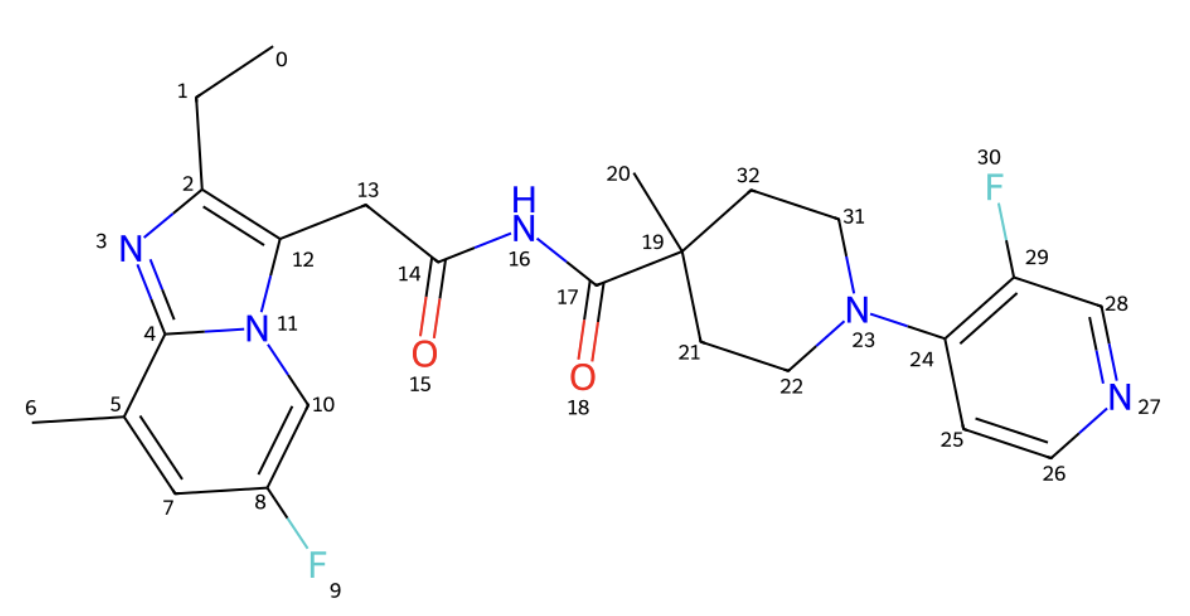}
\end{figure}

In contrast to the MOLLEO generation process, we observe that every modification made to the molecule is perfectly consistent with what the LLM describes in its reasoning trace. We can see that the ligands were cut at the correct desired indices (12 on ligand 1 and 7 on ligand 2) and merged together to form the desired crossover. Then the model describes replacing the chlorine with fluorine to "improve metabolic stability", which is correctly carried out at index 9. Finally, it wishes to add a methyl group to "increase hydrophobic interaction", which is also correctly done at index 5/6. 

\section{Full ToolMol Setup}
\subsection{Toolbox}
\label{app: toolmol_toolbox}
In this section, we provide more detailed descriptions of each tool in the ToolMol toolbox, and specifications for usage of the parameters.
\begin{enumerate}

  \item \texttt{\textbf{add\_atom(mol, idx, element, bond)}} \\
  Adds a single atom of type \texttt{element} to the current molecule. The LLM 
  provides the current \texttt{mol} as a SMILES string. The LLM must also provide 
  the index of the existing atom that will receive a bond to the newly added atom, 
  as well as the type of the bond (i.e. single, double, or triple).

  \item \texttt{\textbf{replace\_atom(mol, idx, element)}} \\
  Replaces the atom at the specified index in the current molecule by a single atom 
  of type \texttt{element}, attempting to preserve all existing bonds.

  \item \texttt{\textbf{add\_functional\_group(mol, idx, group, bond)}} \\
  Adds a single predefined functional group to the current molecule at the specified 
  index. We build a table mapping common functional groups and rings to their SMILES 
  encodings (e.g. methyl, propyl, phenyl, etc), and expose the table to the LLM. It 
  specifies the \texttt{group} parameter in English, and allows the SMILES 
  specification to be handled by the table. Each table entry has a predefined 
  attachment point, and the LLM additionally specifies the bond type for this 
  attachment.

  \item \texttt{\textbf{add\_substructure(mol, idx, substructure, bond)}} \\
  Adds a manually-specified SMILES string substructure to the current molecule at 
  the specified index. Unlike the previous function, the LLM has full flexibility in 
  adding a custom substructure. It must specify an attachment point in 
  \texttt{substructure} in standard \texttt{[*1]} notation, as well as the bond type.

  \item \texttt{\textbf{replace\_substructure(mol, idx, old\_substructure, new\_substructure)}} \\
  Replaces an existing substructure within \texttt{mol} with a new substructure. The 
  LLM specifies \texttt{old\_substructure} using SMARTS (SMILES Arbitrary Target 
  Specification) notation, and provides a new custom substructure as a SMILES. The 
  index parameter here is used as an ``anchor'' to remove ambiguity if the SMARTS 
  matches multiple substructures; only the substructure that contains the \texttt{idx} 
  atom will be selected. Due to the complex nature of removing an entire substructure 
  and reattaching a new one, this function is generally only used on terminal 
  substructures, i.e. substructures that result in only one broken bond when completely 
  removed.

  \item \texttt{\textbf{remove\_substructure(mol, idx, substructure)}} \\
  Removes an entire existing substructure within \texttt{mol}, specified by SMARTS 
  notation. Similar to \texttt{add\_substructure}, the LLM specifies an anchor 
  \texttt{idx} to remove ambiguity in the case of multiple matches. Due to the 
  possibility of creating fragments when removing central substructures, this function 
  is also primarily used on terminal substructures.

  \item \texttt{\textbf{crossover\_molecules(mol1, idx1, mol2, idx2)}} \\
  Takes in two separate molecules as input and performs a crossover operation on them. 
  The LLM specifies 2 indices, one for each molecule. The function then attempts to 
  split both molecules at their respective indices. If successful, this results in 4 
  molecular fragments, from which one of the 4 possible crossover combinations is 
  randomly chosen and returned. This function fails if either index does not result in 
  2 distinct fragments after the split operation (e.g. if the index is part of a ring 
  structure).

\end{enumerate}

\subsection{ToolMol Prompts}
\label{app:toolmol_prompts}
Next, we share detailed information about the prompts and information given to the LLM in ToolMol. First, we introduce two functions that provide important structure and property-based information about an input molecule. These functions are non-callable by the LLM, and are instead deterministically provided before every LLM modification step.

\begin{enumerate}
    \item \texttt{\textbf{get\_ligand\_structure(mol)}}\\
    This function returns the following information for every single atom in the input molecule: \texttt{atom\_index}, \texttt{element}, \texttt{num\_substitutable\_hydrogens}, \texttt{num\_available\_valences}, \texttt{num\_neighboring\_atoms}, \texttt{neighbor\_indices}, \texttt{is\_in\_ring}, and \texttt{centrality}. We observe that the LLM is able to parse this dense atom-wise information quite well, and find this is sufficient to provide the LLM with a solid structural understanding of the ligand. \texttt{centrality} is a measure of how central the atom is with respect to the molecular graph, and is calculated with the betweenness centrality formula. This is an important measure for the LLM to choose a crossover location, as we ideally want to split each molecule as evenly as possible to create the most reasonably-sized fragments. Choosing an atom with high centrality is likely to result in an evenly-split molecule.
    \item \texttt{\textbf{calculate\_properties(mol)}}\\
    This function returns basic molecular descriptors for the input directly provided by RDKit. It returns: \texttt{QED}, \texttt{SA}, \texttt{molecular\_weight}, \texttt{LogP}, \texttt{TPSA}, \texttt{num\_HBond\_donors}, \texttt{num\_HBond\_acceptors}, \texttt{num\_rotatable\_bonds}, and \texttt{num\_aromatic\_rings}.This information primarily helps guide the model's decisions at intermediate steps.
\end{enumerate}

First, below is the system prompt given to the LLM for every modification.
\begin{tcolorbox}[
    breakable,
  colback=brown!8!white,
  colframe=brown!45!white,
  coltitle=brown!60!black,
  colbacktitle=brown!20!white,
  boxrule=0.5pt,
  arc=2pt,
  left=4pt,
  right=4pt,
  top=4pt,
  bottom=4pt,
  title=ToolMol System Prompt
]
You are a molecular design agent.\\
You may ONLY modify molecules using tools.\\
Only make one modification at a time.\\
Read the parameter descriptions for the tools very carefully.\\
Always ensure that your modifications don't break valence rules and do not result in a fragmented molecule.
\end{tcolorbox}

Next is the full initial ToolMol prompt, given to the LLM at the beginning of a modification step.

\begin{tcolorbox}[
  breakable,
  colback=brown!8!white,
  colframe=brown!45!white,
  coltitle=brown!60!black,
  colbacktitle=brown!20!white,
  boxrule=0.5pt,
  arc=2pt,
  left=4pt,
  right=4pt,
  top=4pt,
  bottom=4pt,
  title=Initial ToolMol Prompt
]
"Goal: I want to improve Binding Affinity against [PROTEIN\_TARGET], minimize SA (Synthetic Accessibility), and maximize QED. Recall that a more negative binding affinity is better, and a more positive binding affinity is worse. Please propose a new molecule better than the current molecule. I have given you two candidate ligands. Please propose a new molecule that binds better to [PROTEIN\_TARGET]. You are encouraged to make a crossover between the candidate molecules on the first step, then mutate the resulting molecule. Only make a few modifications (at most 3), then respond with FINAL\_ANSWER. Do not let molecular weight exceed 700.\\\\
1. [LIGAND 1]\\
Binding Affinity against [PROTEIN\_TARGET]: x\\
SA (Synthetic Accessibility): x \\
QED: x \\\\
2. [LIGAND 2]\\
Binding Affinity against [PROTEIN\_TARGET]: x\\
SA (Synthetic Accessibility): x \\
QED: x\\\\
Ligand structure and possible attachment points for ligand 1: [\texttt{get\_ligand\_structure(mol1)}]\\
Ligand structure and possible attachment points for ligand 2: [\texttt{get\_ligand\_structure(mol2)}]\\
Molecule properties for ligand 1: [\texttt{calculate\_properties(mol1)}]\\
Molecule properties for ligand 2: [\texttt{calculate\_properties(mol2)}]
"
\end{tcolorbox}

We outline a multi-objective goal for the LLM, then provide both initial input ligands and the structure / property information using the functions described above. At this initial step, the model chooses to either use the crossover tool to create a new combination, or just uses another tool to modify one of the given input ligands. In either case, one intermediate ligand is produced. We append the tool called and the result to the conversation history.

Following this, we append the following intermediate prompt to the conversation history.

\begin{tcolorbox}[
  colback=brown!8!white,
  colframe=brown!45!white,
  coltitle=brown!60!black,
  colbacktitle=brown!20!white,
  boxrule=0.5pt,
  arc=2pt,
  left=4pt,
  right=4pt,
  top=4pt,
  bottom=4pt,
  title=Intermediate ToolMol Prompts
]
Output FINAL\_ANSWER if you have made sufficient modifications (make at most 3). Ensure that desired properties are maintained. \\
Current SMILES: [CURR\_LIGAND]\\\\
Ligand structure and possible attachment points: [\texttt{get\_ligand\_structure(mol)}]\\
Molecule properties: [\texttt{calculate\_properties(mol)}]\\
\end{tcolorbox}

The model can choose to add additional mutations, and the intermediate prompt is appended to the conversation history after every modification step.

\section{Additional Ablations}
\label{app:additional_ablations}
We provide 3 additional ablations regarding the setup of the genetic algorithm in ToolMol. 

\paragraph{Population \& Offspring Size} We explore using a larger population size of 120 \& and larger offspring size of 70, as well a smaller population size of 12 \& offspring size of 7. This is in contrast to the population size of 60 \& offspring size of 35 used in the ToolMol setup for the main paper.

\paragraph{Pareto Sampling} We also explore an alternate sampling method of choosing parent molecules for crossover and mutation. Instead of sampling proportional to an exponentiated weighted scalar, we consider an approach based on Pareto ordering. Given all molecules $\mathcal{M}_c$ in the current population, we can define multiple Pareto frontiers; let $P_1$ be the set containing the non-dominated frontier on $\mathcal{M}_c$. Then $P_2$ is the set containing the non-dominated Pareto frontier on $\mathcal{M}_c \setminus P_1$, i.e. the next non-dominated frontier obtained after removing all molecules from the true non-dominated frontier from consideration. Then $P_3$ can be defined similarly as the set containing the non-dominated Pareto frontier on $\mathcal{M}_c \setminus (P_1 \cup P_2)$. For this alternate sampling method, we first select the top 3 Pareto frontiers ($P_1$, $P_2$, $P_3$) to proceed to the next generation population after a round of offspring. Then, sampling is determined by each molecule's "rank" in the Pareto ordering. Formally, for a given population of size $n$, the probability of a particular molecule $m_j$ to be selected for crossover / mutation is $P(m_j) = \frac{g(x_j)}{\sum_i g(x_i)}, i \in \{1, ..., n\}$, where $x_i = \{1 \text{ if } m_i \in P_1 \text{, } 2 \text{ if } m_i \in P_2\text{, } 3 \text{ if } m_i \in P_3$\}, and $g(x) = \frac{1}{1 + x}$. We consider this sampling method because it aligns well with the Pareto approach we take to multi-objective optimization in the rest of the genetic algorithm. 

Table \ref{ablation3} shows the results of the three aforementioned ablations, compared against the ToolMol setup shown in the main paper. We run all ablations on 3 different seeded initial populations.

\begin{table}[H]
\centering
\caption{Additional Ablations on ToolMol GA: Population Size \& Pareto-rank Sampling}
\label{ablation3}
\small
\resizebox{0.6\textheight}{!}{%
\begin{tabular}{ll|cccc}
\toprule
Target & Metric & \makecell{ToolMol \\ (12 / 7)} & \makecell{ToolMol \\ (120 / 70)} & \makecell{ToolMol \\ (Pareto Sampling)} & ToolMol \\
\midrule
\multirow{3}{*}{c-MET}
 & Binding Affinity ($\downarrow$)  & $\mathbf{-11.14 \pm 0.20}$ & $-10.68 \pm 0.12$ & $\underline{-11.07 \pm 0.19}$ & $-11.00 \pm 0.09$ \\
 & Filtered Affinity ($\downarrow$) & $-10.22 \pm 0.16$ & $-10.13 \pm 0.09$  & $\underline{-10.27 \pm 0.06}$ & $\mathbf{-10.35 \pm 0.17}$ \\
 & Hypervolume ($\uparrow$)                     & $\underline{0.60 \pm 0.02}$ & $0.59 \pm 0.01$ & $\underline{0.60 \pm 0.01}$ & $\mathbf{0.62 \pm 0.01}$   \\
\midrule
\multirow{3}{*}{BRD4}
 & Binding Affinity ($\downarrow$)  & $-10.61 \pm 0.33$ & $\underline{-10.79 \pm 0.23}$ & $\mathbf{-10.95 \pm 0.14}$ & $-10.64 \pm 0.28$ \\
 & Filtered Affinity ($\downarrow$) & $\underline{-9.87 \pm 0.32}$ & $\underline{-9.87 \pm 0.29}$ & $-9.67 \pm 0.19$ & $\mathbf{-9.91 \pm 0.18}$  \\
 & Hypervolume ($\uparrow$)           & $0.59 \pm 0.01$ & $\mathbf{0.60 \pm 0.02}$  & $0.59 \pm 0.004$ & $\mathbf{0.60 \pm 0.01}$   \\
\midrule
\multirow{3}{*}{ACAA1}
 & Binding Affinity ($\downarrow$)  & $\mathbf{-9.87 \pm 0.18}$ & $-9.54 \pm 0.08$ & $\mathbf{-9.87 \pm 0.19}$ & $-9.70 \pm 0.23$  \\
 & Filtered Affinity ($\downarrow$) & $-8.77 \pm 0.24$ & $\mathbf{-8.86 \pm 0.25}$ & $\underline{-8.81 \pm 0.32}$ & $-8.78 \pm 0.15$  \\
 & Hypervolume ($\uparrow$)         & $0.53 \pm 0.02$ & $\mathbf{0.54 \pm 0.001}$  & $\mathbf{0.54 \pm 0.009}$ & $\mathbf{0.54 \pm 0.008}$   \\
\midrule
 & Avg. Rank ($\downarrow$)         & $2.67$ & $2.56$ & $\underline{2.00}$ & $\mathbf{1.89}$ \\
\bottomrule
\end{tabular}
}
\label{tab:results}
\end{table}

We observe that the configuration of ToolMol in the main paper beats all aforementioned ablations on average across all targets. Thus, we choose to report the values and setup of the rightmost column in comparison to other baselines in our main analysis, although it is likely that the Pareto sampling method is not significantly weaker, and even beats the exponential fitness setup on particular targets.

We also briefly test an alternate value for the exponential constant $k$ used in parent sampling. We use $k=10$ in the main paper, and test that against $k=e$ here for the c-MET target. Results are shown in Table \ref{ablation4}.
\begin{table}[H]
\centering
\caption{Ablation on exponential constant $k$: $10$ vs $e$}
\label{ablation4}
\small
\resizebox{0.4\textheight}{!}{%
\begin{tabular}{ll|cc}
\toprule
Target & Metric & $k=e$ & $k=10$ \\
\midrule
\multirow{3}{*}{c-MET}
 & Binding Affinity ($\downarrow$)  & $\mathbf{-11.12 \pm 0.18}$ & $-11.00 \pm 0.09$ \\
 & Filtered Affinity ($\downarrow$) & $\-10.32 \pm 0.15$ & $\mathbf{-10.35 \pm 0.17}$ \\
 & Hypervolume ($\uparrow$)         & $0.61 \pm 0.01$   & $\mathbf{0.62 \pm 0.01}$ \\
\bottomrule
\end{tabular}
}
\label{tab:results}
\end{table}

We observe that while there is very little difference between the 2 constants, using $k=10$ marginally improves the multi-objective metrics we care most about, and thus we choose to report that configuration in the main paper.

\section{AlphaEvolve / ShinkaEvolve for Drug Discovery}
\label{app: shinkaevolve}

In this section, we outline the ShinkaEvolve-inspired algorithm we built for small-molecule drug discovery. It is a non-sophisticated MAP-Elites approach with independent islands and random migration events. We maintain 4 separate MAP-Elites grids that are referred to as islands. Each grid is actually 1-dimensional, and stores molecule candidates within bins based on their molecular weight. There are 50 bins evenly discretizing molecular weight within the range [200, 900]. Any molecules outside of that range are placed into the outermost bins. 

The core LLM modification step occurs when we sample two molecules from a particular grid for crossover / mutation operations, resulting in one new molecule. The sampling procedure closely follows ShinkaEvolve, where parent molecules are sampled based on a balance between fitness and how often that molecule has already been sampled for reproduction. Let $\Phi(m) = \sum_if_i(m)$ be the fitness of a molecule, where $f_i(m)$ is the $i$th objective scaled to $[0, 1]$. Let $\alpha = \text{median}(\Phi(m_1), ..., \Phi(m_n))$ for all $m_i$ currently in the MAP-Elites grid. Then let $s_i = \sigma(\lambda*(\Phi(m_i) - \alpha))$, where $\sigma$ is the sigmoid function and $\lambda$ is a constant that controls selection pressure. We use $\lambda = 1$ for our experiments. Further, let $h_i = \frac{1}{1 + N(m_i)}$, where $N(m_i)$ counts the number of times $m_i$ has already been chosen for reproduction. Thus for each molecule $m_i$, we have $s_i$ which benefits molecules with high fitness, and $h_i$ which benefits molecules that have not been chosen frequently. The final probability distribution is constructed by $P(m_i) = \frac{w_i}{\sum_jw_j}, j \in \{1, ..., n\}$, where $w_i = s_i * h_i$. After 2 molecules are selected according to this sampling formulation, they undergo LLM crossover / mutation steps either in a similar manner to MOLLEO, or with the ToolMol toolbox.

When a new generated molecule is trying to get placed into the grid, if the bin corresponding to the new molecule is not filled, the new molecule is immediately placed into that bin. If it is occupied, the new molecule replaces the current molecule in the bin only if it has a higher fitness, calculated by the $\Phi(m)$ formula described above.

On initialization of the algorithm, we sample 40 molecules from ZINC 250K, and place them uniformly at random across the 4 islands. Then, each island undergoes 10 independent molecule generations. After all islands complete their generations, a migration event occurs; we sample 2 molecules from each island uniformly at random, then send a copy of those molecules to another island, also chosen uniformly. Whether or not those migrated molecules are accepted into the island depends on the bin they land into and the fitness competition described above. Following ShinkaEvolve, we do not allow the absolute highest fitness molecule from each island to migrate, aiming to preserve some level of diversity between the islands. After 1000 total binding affinity oracle evaluations, the algorithm terminates, and all generated molecules (including ones discarded due to losing to fitness competition) are returned for downstream evaluation. 

We do not implement the novelty rejection-sampling or the LLM ensemble described in ShinkaEvolve for our simplistic implementation. We note that this algorithm is still largely unexplored for drug discovery problems, and anticipate that there are likely significant gains to be made beyond our simplistic implementation that was designed primarily as a baseline. We plan to explore further variations of this algorithm for this multi-objective problem in future work.

\section{Additional ABFE Information}
\label{app:abfe_setup}
\subsection{ABFE Setup}

For our ABFE calculations, we utilize the following Binding Affinity Tool \texttt{BAT.py} \citep{Heinzelmann2021} repository: \url{https://github.com/GHeinzelmann/BAT.py}. We simulate using OpenMM and the standard SDR method. We use the Boltz-2 predicted ligand pose as the starting pose for the simulation. Because Boltz-2 does not take a protein crystal structure as input and makes a prediction based on the given amino acid sequence, we first align the entire predicted Boltz-2 conformation to the protein crystal structure with ChimeraX \citep{Pettersen2020}, then extract only the ligand pose for ABFE. We observe this alignment to yield an RMSE of under 0.7 angstroms on average; thus we are comfortable using the aligned ligand pose with the crystal structure in ABFE calculations. We do not observe frequent steric clashes resulting from this process.

Our simulation steps parameters for the BAT.py framework are as follows:\\
eq\_steps1 = 500000 (Number of steps for equilibration gradual release)\\
eq\_steps2 = 15000000 (Number of steps for equilibration after release)
 
m\_steps1 = 500000 (Number of steps per window for component m (equilibrium))\\
m\_steps2 = 1000000 (Number of steps per window for component m (production))

n\_steps1 = 500000 (Number of steps per window for component n (equilibrium))\\
n\_steps2 = 1000000 (Number of steps per window for component n (production))

e\_steps1 = 250000 (Number of steps per window for component e (equilibrium))\\
e\_steps2 = 500000 (Number of steps per window for component e (production))

v\_steps1 = 500000 (Number of steps per window for component v (equilibrium))\\
v\_steps2 = 1000000 (Number of steps per window for component v (production))

On 8 NVIDIA RTX 4090 GPUs, one ABFE calculation typically takes around 12 hours to complete.

\subsection{Correlation Analysis: ABFE vs Boltz-2 vs AutoDock}
\label{correlation_analysis}
\begin{figure}[htbp]
    \centering
    \begin{subfigure}{0.48\textwidth}
        \centering
        \includegraphics[width=\textwidth]{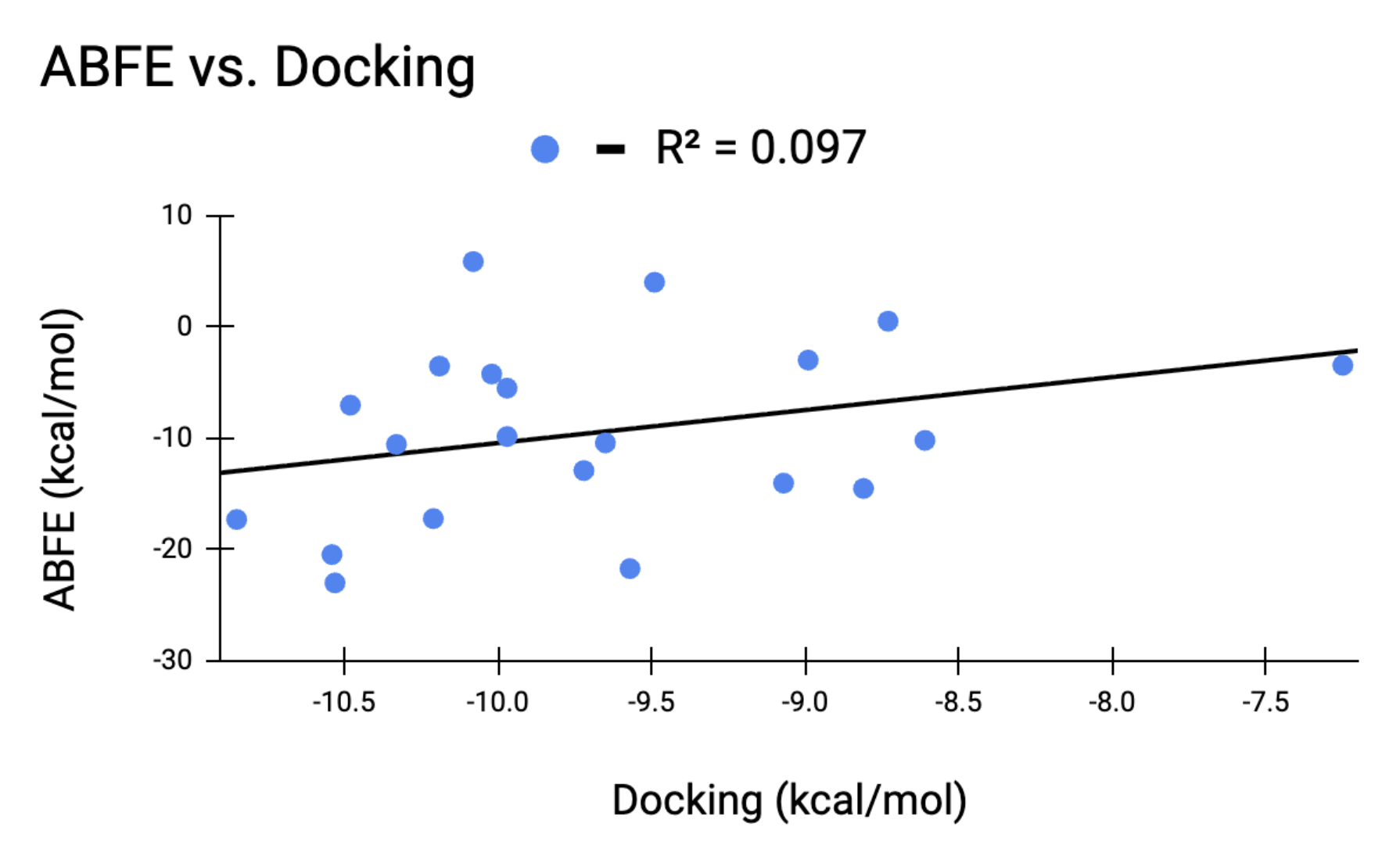}
        \caption{ABFE vs AutoDock scores}
    \end{subfigure}
    \hfill
    \begin{subfigure}{0.48\textwidth}
        \centering
        \includegraphics[width=\textwidth]{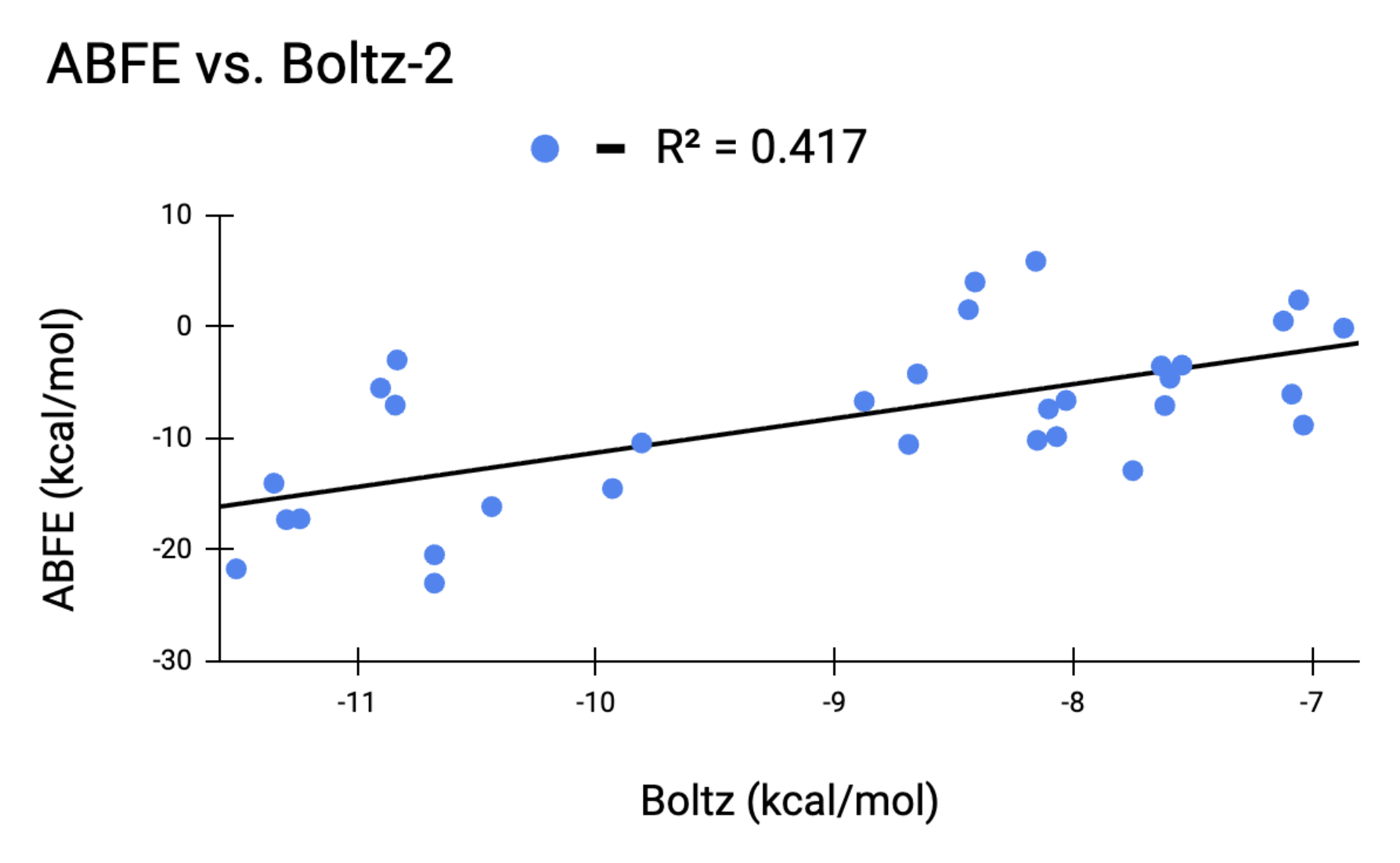}
        \caption{ABFE vs Boltz-2 scores}
    \end{subfigure}
    \caption{Comparison of correlation between AutoDock \& ABFE and Boltz-2 \& ABFE for 32 known compounds for the c-MET protein target. We observe a significantly higher correlation between Boltz-2 and ABFE as compared to AutoDock.}
    \label{fig abfe correlation}
\end{figure}

To justify our usage of Boltz-2 as a primary binding affinity oracle, we provide general analysis of the correlation between Boltz-2, AutoDock, and the gold-standard ABFE. In Figure \ref{fig abfe correlation}, we take 32 compounds for c-MET, 16 of which are known binders, and 16 of which are presumed inactive binders. We calculate the ABFE, Boltz-2, and AutoDock binding affinities for all 32 compounds. We exclude results for any failed AutoDock or Boltz-2 runs.

We see that ABFE and AutoDock docking show $r^2 = 0.09$ among the 32 compounds, while ABFE and Boltz-2 show $r^2 = 0.42$. As an oracle nearly 1000x less computationally expensive than ABFE, Boltz-2 shows exceptional correlation with ABFE, especially in comparison to docking. Furthermore, we calculate the ROC-AUC score for Boltz-2 and docking, to see how well they can separate binders from non-binders. Boltz-2 scores 0.95 for this metric, while AutoDock scores 0.84. Due to computational and time constraints regarding expensive ABFE calculations, we are only able to provide results for the c-MET target at this time.

We demonstrate that Boltz-2 has stronger correlation with the most accurate gold-standard computational methods for one of our primary binding targets, which motivates us to employ Boltz-2 as a binding affinity oracle over the current industry-standard AutoDock, which itself has often been noted for its practical inaccuracy. We generally observe Boltz-2 to be approximately a factor of 10 more expensive to run than AutoDock; however, this difference is entirely negligible in comparison to the cost of molecular dynamics methods such as ABFE.

\end{document}